\documentclass[runningheads]{llncs}
\usepackage[T1]{fontenc}
\usepackage{graphicx}
\usepackage{mathrsfs}
\usepackage{multirow}%
\usepackage{amsmath,amssymb,amsfonts}%
\usepackage[title]{appendix}%
\usepackage{xcolor}%
\usepackage{textcomp}%
\usepackage{manyfoot}%
\usepackage{booktabs}%
\usepackage{algorithm}%
\usepackage{algorithmicx}%
\usepackage{algpseudocode}%
\usepackage{listings}%

\usepackage[utf8]{inputenc}
\usepackage[T1]{fontenc}

\usepackage{graphicx}
\usepackage{svg}
\usepackage{tikz}

\usepackage{amsmath,amssymb,amsfonts}%

\usepackage{algorithm}%
\usepackage{algorithmicx}%
\usepackage{algpseudocode}%
\usepackage{cleveref}
\usepackage{pbox}
\usepackage{makecell}
\usepackage{soul}
\usepackage{xcolor}

\usepackage[numbers]{natbib}

\begin{document}
\pagestyle{empty}  
\title{Evaluating the Effectiveness of SechKAN on 1D Data}
%
%
\author{
Hoang-Thang Ta\inst{1,2}\orcidID{0000-0003-0321-5106}
} 
%
\authorrunning{Ta et al.}
%
\institute{
University of Information Technology, Ho Chi Minh City, Vietnam\\
\and
Vietnam National University Ho Chi Minh City, Ho Chi Minh City, Vietnam\\
\email{thangth@uit.edu.vn}
}
%
\maketitle              
\begin{abstract}
The connection between the Kolmogorov-Arnold representation theorem (KART) and neural network design has led to the development of Kolmogorov-Arnold Networks (KANs), with applications ranging from STEM problems to AI tasks. In this paper, we investigate the effectiveness of a KAN variant, SechKAN, which relies on hyperbolic secant (sech) functions as basis functions, with a 1D projection to reduce the number of parameters to a level comparable to MLPs. We evaluate SechKAN on three 1D classification datasets: UCI Human Activity Recognition (UCI HAR), ElectricDevices, and Crop, and compare it with several effective networks, including EfficientKAN, MLP, CNN1D, ResNet1D, and DSCNN1D, using approximately comparable parameter budgets. The results indicate that SechKAN achieves competitive performance across the three datasets, with particularly strong performance on Crop. Ablation studies further show that grid size and normalization affect performance, suggesting that SechKAN's effectiveness depends on the dataset and architectural choices. Our source code and experimental implementation are publicly available at: \url{https://github.com/hoangthangta/SechKAN\_1D}.

\keywords{Kolmogorov-Arnold Networks \and SechKAN \and Hyperbolic Secant Functions \and One-Dimensional Data}
\end{abstract}
\section{Introduction}
Neural networks have achieved significant success in developing models that learn from data and mimic aspects of human intelligence. Neural network architectures have evolved from early architectures such as MLPs to RNNs, CNNs, LSTMs, TCNs, and many new architectures, especially LLMs, and have been applied to a wide range of tasks. However, while MLPs are simple and widely used due to their effectiveness, their linear transformations cannot always adapt well to different types of data. The introduction of KANs has provided an alternative approach for many tasks involving functions and relationships in mathematics, physics, and other fields~\cite{somvanshi2025survey}.

Continuing research on novel KAN architectures, we investigate SechKAN, a KAN that uses hyperbolic secant functions as basis functions. Unlike other KANs, SechKAN applies a 1D projection after feature extraction using the sech functions to reduce the number of parameters in its layers, making them comparable to those of MLP layers. In previous work, SechKAN showed its effectiveness on several tasks, including function approximation, PDE surrogate modeling, and image classification~\cite{ta2026sechkan}. However, its performance on 1D data remains unexplored, despite its design for one-dimensional inputs.

An evaluation of SechKAN on 1D classification data is still lacking, particularly across diverse datasets and architectural settings. To address this gap, we evaluate SechKAN on three benchmark datasets: UCI HAR, ElectricDevices, and Crop. We compare it with EfficientKAN, MLP, CNN1D, ResNet1D, and DSCNN1D, covering KAN-based, fully connected, and convolutional architectures. EfficientKAN is included as a representative KAN baseline based on its competitive performance in our previous comparison of KAN variants~\cite{ta2026sechkan}, while the CNN-based models are selected for their ability to extract local patterns from one-dimensional data. We also conduct ablation studies on grid size (number of basis functions) and data normalization. The main contribution of this work is a systematic empirical evaluation of SechKAN across three 1D classification datasets, including comparisons with KAN-, MLP-, and CNN-based models, as well as ablation studies of key settings.

The remainder of this paper is organized as follows. Section 2 introduces related work to our SechKAN, including research directions related to KAN and KART. Section 3 introduces the SechKAN architecture and other KAN variants, as well as CNN-based models. Section 4 contains experimental information, including datasets, training configurations, evaluation metrics, experimental results, and ablation studies. Section 5 presents the limitations of the study. Finally, Section 6 concludes the paper and discusses future work.

\section{Related Work} 
KANs were recently introduced by \citet{liu2025kan} as a new approach to neural network design. Unlike MLPs, KANs use learnable univariate functions instead of fixed activation functions. KANs are based on the KART, which states that a continuous multivariate function can be represented using a finite combination of one-dimensional functions~\cite{kolmogorov1957representation,braun2009constructive}. Before KANs, KART had already inspired several neural network architectures, including spline-based models~\cite{leni2013kolmogorov,van2022kasam}.

KANs have been applied to various tasks, including differential equation solving~\cite{wang2025kolmogorov}, time series forecasting~\cite{genet2024temporal}, and computer vision~\cite{li2025u}. Many studies have also explored alternative basis functions to replace the original B-spline basis. These include modified B-splines~\cite{ta2024bsrbf}, polynomial functions~\cite{ss2024chebyshev}, radial basis functions~\cite{li2024kolmogorov}, Fourier bases~\cite{xu2024fourierkan}, wavelets~\cite{bozorgasl2024wav}, rational and fractional Jacobi functions~\cite{aghaei2026rkan,aghaei2025fkan}, and customized activation functions~\cite{qiu2025relu}.

Another research direction focuses on reducing the large number of parameters in KANs. \citet{yang2025kolmogorov} introduced Group KAN in Kolmogorov-Arnold Transformers (KATs), which reduces the number of parameters and computational costs through weight sharing. \citet{ta2025prkan} proposed PRKAN, which uses several parameter-reduction techniques to achieve a model size comparable to MLPs. Inspired by PRKAN, SechKAN uses hyperbolic secant functions as basis functions and a 1D projection to reduce the number of parameters~\cite{ta2026sechkan}. Other parameter-efficient KANs include GS-KAN, which generates edge functions from a shared learnable parent function~\cite{eliasson2025gs}, LeanKAN, which provides a compact alternative to AddKAN and MultKAN~\cite{koenig2025leankan}.

CNN-based architectures are widely used for data with local and sequential structures. For 1D data, CNN1D applies one-dimensional convolution to extract local features from input sequences~\cite{kiranyaz20211d}. ResNet1D extends this approach with residual connections, enabling deeper networks to learn effectively~\cite{he2016deep}. DSCNN1D uses depthwise separable convolutions to reduce computational cost and the number of parameters while preserving local feature extraction capabilities~\cite{sorensen2020depthwise,majumdar2020matchboxnet}. These architectures provide established convolutional baselines for evaluating SechKAN on 1D classification tasks.

\section{Methodology}

\subsection{KART and KAN}
Kolmogorov--Arnold Networks (KANs) are based on the Kolmogorov--Arnold Representation Theorem (KART), which states that any continuous multivariate function on a bounded domain can be represented using a finite sum of one-dimensional functions. Let $\mathbf{x} = (x_1, \ldots, x_n) \in [0,1]^n$. Any continuous function $f : [0,1]^n \to \mathbb{R}$ can be represented as~\cite{ismailov2025addressing}:

\begin{equation}
\begin{aligned}
f(\mathbf{x}) = f(x_1, \ldots, x_n) = \sum_{q=1}^{2n+1} \Phi_q \left( \sum_{p=1}^{n} \phi_{q,p}(x_p) \right)
\end{aligned}
\label{eq:kart}
\end{equation}

Here, $\phi_{q,p}$ are one-dimensional functions applied to individual input variables, while $\Phi_q$ combines their outputs. This decomposition provides the theoretical basis for representing multivariate functions using one-dimensional functions.

\citet{liu2025kan} introduced KANs as a neural network architecture based on this theorem. Instead of using fixed activation functions and linear weights as in MLPs, KANs use learnable one-dimensional functions on the edges. A KAN with $L$ layers applies a sequence of function matrices $\Phi_0, \Phi_1, \ldots, \Phi_{L-1}$:

\begin{equation}
\begin{aligned}
\text{KAN}(\mathbf{x}) = (\Phi_{L-1} \circ \Phi_{L-2} \circ \cdots \circ \Phi_1 \circ \Phi_0)\mathbf{x}
\end{aligned}
\label{eq:kan}
\end{equation}

For a layer $l$, the function matrix $\Phi_l \in \mathbb{R}^{n_{l+1}\times n_l}$ contains learnable functions $\phi_{l,j,i}$ connecting the $i$-th node of layer $l$ to the $j$-th node of layer $l+1$:

\begin{equation}
\begin{aligned}
\phi_{l,j,i}, \quad l = 0, \cdots, L - 1, \quad i = 1, \cdots, n_l, \quad j = 1, \cdots, n_{l+1}
\end{aligned}
\label{eq:acti_funct}
\end{equation}

With \(n_l\) nodes in the \(l^{th}\) layer, the transformation from \(\mathbf{x}_l\) to \(\mathbf{x}_{l+1}\) is computed by the function matrix \(\Phi_l \in \mathbb{R}^{n_{l+1}\times n_l}\):  

\begin{equation}
\begin{aligned}
\mathbf{x}_{l+1} = 
\underbrace{\left(
\begin{array}{cccc}
\phi_{l,1,1}(\cdot) & \phi_{l,1,2}(\cdot) & \cdots & \phi_{l,1,n_l}(\cdot) \\
\phi_{l,2,1}(\cdot) & \phi_{l,2,2}(\cdot) & \cdots & \phi_{l,2,n_l}(\cdot) \\
\vdots & \vdots & \ddots & \vdots \\
\phi_{l,n_{l+1},1}(\cdot) & \phi_{l,n_{l+1},2}(\cdot) & \cdots & \phi_{l,n_{l+1},n_l}(\cdot)
\end{array}\right)}_{\Phi_{l}} \mathbf{x}_l
\label{eq:function_matrix}
\end{aligned}
\end{equation}

In the original KAN, each learnable function is constructed using a base function and a spline function~\cite{liu2025kan}:

\begin{equation}
\begin{aligned}
\phi(x) = w_b b(x) + w_s spline(x)
\end{aligned}
\label{eq:acti_funct_imp}
\end{equation}

where the base function is the SiLU function:

\begin{equation}
\begin{aligned}
b(x) = silu(x) = \frac{x}{1 + e^{-x}}
\end{aligned}
\label{eq:b_function}
\end{equation}

and the spline function is defined as:

\begin{equation}
\begin{aligned}
spline(x) = \sum_{i}c_iB_i(x)
\end{aligned}
\label{eq:spline_function}
\end{equation}

where $c_i$ are learnable coefficients and $B_i(x)$ are B-spline basis functions.

\subsection{SechKAN}

\begin{figure*}[!ht]
  \centering
  \includegraphics[scale=0.71]{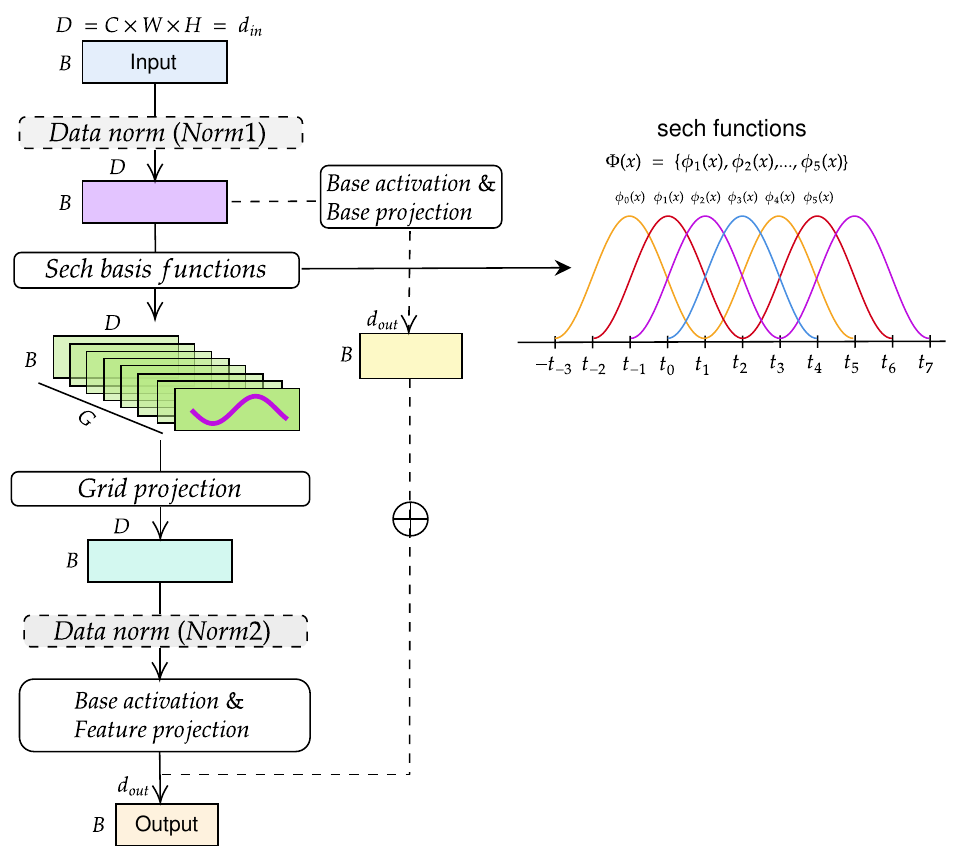}
\caption{Architecture of a SechKAN layer, consisting of sech basis functions, grid projection, normalization, feature projection, and a parallel base-projection branch. The figure is adapted from the SechKAN work~\cite{ta2026sechkan}. Note that the illustrated sech basis functions are simulated for visualization purposes; they asymptotically approach zero as the input moves away from the center but remain nonzero for any finite input.}
  \label{fig:sechkan_layer}
\end{figure*}

SechKAN consists of multiple layers, where each layer processes the output of the previous layer, with sech functions serving as the main feature extractor and a grid projection acting as a 1D projection to reduce the number of parameters~\cite{ta2026sechkan}. As illustrated in ~\Cref{fig:sechkan_layer}, an input first passes through an optional Norm1, followed by sech basis functions that transform each input feature into \(G\) basis responses (the number of basis functions), introducing an additional grid dimension. The grid projection then aggregates the \(G\) basis responses into a single representation for each input feature. The resulting representation can pass through an optional second normalization position (Norm2), followed by a base activation and feature projection to produce the main output. In parallel, a skip projection branch processes the normalized input using the same base activation followed by a linear projection. Its output is then added element-wise to the main output to produce the final output. If the skip projection branch is disabled, the layer directly returns the main output.

The sech function is smooth, bell-shaped, exponentially decaying, and infinitely differentiable, supporting smooth gradient propagation and localized representations. Its bounded output can help maintain stable activations, although saturation may lead to small gradients for inputs far from the center. These properties make sech functions suitable as basis functions for grid-based feature representations in SechKAN, while sharing characteristics with other localized basis functions used in KANs.

For a SechKAN layer, the input \(X\) has shape \((B, D)\), where \(B\) is the batch size and \(D\) is the data dimension. For image inputs, \(D=C\times W\times H\), where \(C\), \(W\), and \(H\) denote the number of channels, width, and height, respectively. We flatten \(X\) to obtain a tensor of shape \((B,d_{\mathrm{in}})\), where \(d_{\mathrm{in}}\) is the input dimension. The output \(Y\) has shape \((B,d_{\mathrm{out}})\), where \(d_{\mathrm{out}}\) is the output dimension.

Suppose that \(X'\) denotes the input after applying Norm1. We then map \(X'\) to a set of sech basis functions. Let \(G\) denote the number of basis functions (grid size), and let \(\mathcal{C}=\{c_g\}_{g=1}^{G}\) denote their learnable center locations. The centers are initialized uniformly over the interval \([g_{\min},g_{\max}]\) as

\begin{equation}
c_g =
g_{\min}+
\frac{g-1}{G-1}(g_{\max}-g_{\min}),
\quad g=1,\ldots,G.
\end{equation}

During training, the center locations are jointly optimized with the other model parameters, while the number of basis functions \(G\) remains fixed.

The sech basis responses are defined as

\begin{equation}
\Phi(X') =
s\cdot
\frac{1}
{\cosh\!\left(\frac{X'-\mathcal{C}}{w}\right)}
+\delta,
\quad
\Phi(X')\in
\mathbb{R}^{B\times d_{\mathrm{in}}\times G},
\label{eq:sech_basis}
\end{equation}
where \(s\) and \(\delta\) are learnable scale and bias parameters, respectively, and \(w\) is a learnable width parameter shared by all sech basis functions within a layer. The width is parameterized through an unconstrained variable \(\theta_w\) as

\begin{equation}
w=\exp(\theta_w)+10^{-8},
\end{equation}
which ensures a positive width and improves numerical stability during optimization.





\section{Experiments}
\subsection{Datasets and Training Configurations}

Three benchmark one-dimensional classification datasets were used: UCI Human Activity Recognition (UCI HAR), ElectricDevices, and Crop.

\begin{itemize}
    \item \textbf{UCI HAR.} The dataset contains 10,299 samples represented by 561-dimensional feature vectors across six human activities, with 7,352 training and 2,947 test samples in the original subject-independent split~\cite{anguita2013public}. The test set was kept unchanged, while 20\% of the training subjects were held out for validation using a fixed random seed of 42, ensuring subject-disjoint training and validation sets.
    
    \item \textbf{ElectricDevices.} The dataset contains 16,637 univariate sequences of length 96 from seven classes, with 8,926 training and 7,711 test samples.\footnote{\url{https://www.timeseriesclassification.com/description.php?Dataset=ElectricDevices}} The test set was kept unchanged, while the training set was stratified by class into 80\% training and 20\% validation samples using a fixed random seed of 42.
    
    \item \textbf{Crop.} The dataset contains 24,000 sequences of length 46 from 24 classes, with 7,200 training and 16,800 test samples.\footnote{\url{https://www.timeseriesclassification.com/description.php?Dataset=Crop}} The test set was kept unchanged, while the training set was stratified by class into 80\% training and 20\% validation samples using a fixed random seed of 42. 
\end{itemize}

All models were trained on an NVIDIA GeForce RTX 3060 Ti GPU using AdamW with a learning rate of $10^{-3}$, weight decay of $10^{-4}$, batch size of 64, and a OneCycleLR scheduler. The data splits were fixed using random seed 42, while model training was independently repeated five times using random seeds $0$--$4$. The reported results are presented as mean $\pm$ standard deviation across five runs (using five seeds) in the main experiments and three runs (using three seeds) in the ablation studies.

The architectures were designed with approximately comparable numbers of trainable parameters for fair comparison (\Cref{tab:model_architectures}). The MLP used a single hidden layer of 256 units with LayerNorm and SiLU. CNN1D, ResNet1D, and DSCNN1D served as 1D convolutional baselines, using three convolutional layers, an initial convolution followed by three residual blocks, and six depthwise-separable convolutional blocks, respectively. All three models used BatchNorm and SiLU, followed by global average pooling and a fully connected output layer. EfficientKAN used 256 hidden units, a grid size of 5, and spline order 3.

SechKAN used a 256-unit hidden layer, with 4 basis functions (or grid size=4) and LayerNorm for UCI HAR and ElectricDevices, and 16 basis functions (or grid size = 16) with BatchNorm for Crop. The second normalization stage was disabled, while SiLU was used as the base activation and the base-function update was disabled. The learnable width was disabled for UCI HAR and Crop but enabled for ElectricDevices. Models were trained for 20 epochs on UCI HAR and Crop and 30 epochs on ElectricDevices. The exact network structures and trainable parameter counts are reported in ~\Cref{tab:model_architectures}.

\subsection{Results}

\begin{table*}[ht]
	\caption{Network structures and numbers of trainable parameters for the models on the three datasets.}
	\centering
	\resizebox{\textwidth}{!}{%
	\begin{tabular}{p{1.8cm}p{2.1cm}p{1.7cm}p{7.0cm}}
            \hline
		\textbf{Dataset} & \textbf{Model} & \textbf{Params} & \textbf{Network structure} \\
            \hline
		\multirow{6}{1.8cm}{\textbf{UCI HAR}}
		& CNN1D & 147,048 & 3-layer 1D CNN. \\
		& DSCNN1D & 147,038 & 6 depthwise-separable convolutional blocks. \\
		& EfficientKAN & 147,420 & $561$--$256$--$6$; grid size $=5$, spline order $=3$. \\
		& MLP & 147,048 & $561$--$256$--$6$; LayerNorm and SiLU. \\
		& ResNet1D & 147,048 & Initial convolution + 3 residual blocks. \\
		& SechKAN & 147,070 & $561$--$256$--$6$; grid size = 4, LayerNorm, SiLU. \\
            \hline
		\multirow{6}{1.8cm}{\textbf{Electric Devices}}
		& CNN1D & 27,335 & 3-layer 1D CNN. \\
		& DSCNN1D & 27,335 & 6 depthwise-separable convolutional blocks. \\
		& EfficientKAN & 26,780 & $96$--$256$--$7$; grid size $=5$, spline order $=3$. \\
		& MLP & 27,335 & $96$--$256$--$7$; LayerNorm and SiLU. \\
		& ResNet1D & 27,335 & Initial convolution + 3 residual blocks. \\
		& SechKAN & 27,359 & $96$--$256$--$7$; grid size = 4, LayerNorm, SiLU; learnable width enabled. \\
            \hline
		\multirow{6}{1.8cm}{\textbf{Crop}}
		& CNN1D & 18,804 & 3-layer 1D CNN. \\
		& DSCNN1D & 18,804 & 6 depthwise-separable convolutional blocks. \\
		& EfficientKAN & 18,200 & $46$--$256$--$24$; grid size $=5$, spline order $=3$. \\
		& MLP & 18,804 & $46$--$256$--$24$; LayerNorm and SiLU. \\
		& ResNet1D & 18,804 & Initial convolution + 3 residual blocks. \\
		& SechKAN & 18,874 & $46$--$256$--$24$; grid size = 16, BatchNorm, SiLU. \\
            \hline
		\multicolumn{4}{l}{Params = number of trainable parameters.} \\
            \hline
	\end{tabular}%
	}
	\label{tab:model_architectures}
\end{table*}

\begin{table*}[ht]
	\caption{Performance comparison of different models on UCI HAR, ElectricDevices, and Crop. Values are reported as mean $\pm$ standard deviation over five independent training runs using seeds 0--4.}
	\centering
	\begin{tabular}{p{1.8cm}p{2.5cm}p{2.5cm}p{2.5cm}p{2.2cm}}
            \hline
		\textbf{Dataset} & \textbf{Model} & \textbf{Val. Acc.} & \textbf{Test Acc.} & \textbf{Time (s)} \\
            \hline
		\multirow{6}{2cm}{\textbf{UCI HAR}}
		& CNN1D & 93.49 $\pm$ 1.77 & 94.94 $\pm$ 0.22 & 9.68 $\pm$ 0.48 \\
		& DSCNN1D & 92.35 $\pm$ 2.35 & 94.03 $\pm$ 1.08 & 19.85 $\pm$ 3.83 \\
		& EfficientKAN & 92.98 $\pm$ 2.36 & \textbf{94.97 $\pm$ 0.54} & 10.80 $\pm$ 0.21 \\
		& MLP & 92.98 $\pm$ 2.46 & 94.20 $\pm$ 1.15 & \textbf{6.93 $\pm$ 0.80} \\
		& ResNet1D & 92.73 $\pm$ 2.81 & 94.05 $\pm$ 0.78 & 14.72 $\pm$ 1.25 \\
		& SechKAN & 93.22 $\pm$ 2.52 & 94.84 $\pm$ 0.86 & 9.74 $\pm$ 0.79 \\
            \hline
		\multirow{6}{2cm}{\textbf{Electric Devices}}
		& CNN1D & 76.82 $\pm$ 0.64 & 58.66 $\pm$ 1.32 & 21.53 $\pm$ 2.08 \\
		& DSCNN1D & 81.04 $\pm$ 1.11 & 62.43 $\pm$ 1.10 & 36.51 $\pm$ 1.18 \\
		& EfficientKAN & 75.00 $\pm$ 0.41 & 65.69 $\pm$ 0.42 & \textbf{16.75 $\pm$ 1.64} \\
		& MLP & 73.71 $\pm$ 0.27 & 56.30 $\pm$ 0.73 & 15.02 $\pm$ 0.27 \\
		& ResNet1D & \textbf{87.60 $\pm$ 0.25} & \textbf{69.86 $\pm$ 0.61} & 25.89 $\pm$ 0.93 \\
		& SechKAN & 75.52 $\pm$ 1.09 & 62.75 $\pm$ 0.85 & 20.24 $\pm$ 0.41 \\
            \hline
		\multirow{6}{2cm}{\textbf{Crop}}
		& CNN1D & 71.39 $\pm$ 0.58 & 72.30 $\pm$ 0.23 & 11.23 $\pm$ 1.88 \\
		& DSCNN1D & 72.31 $\pm$ 1.00 & 73.03 $\pm$ 0.51 & 20.61 $\pm$ 4.10 \\
		& EfficientKAN & 62.42 $\pm$ 0.32 & 62.82 $\pm$ 0.09 & 12.15 $\pm$ 0.80 \\
		& MLP & 68.78 $\pm$ 0.31 & 68.76 $\pm$ 0.24 & \textbf{7.61 $\pm$ 0.43} \\
		& ResNet1D & 71.42 $\pm$ 0.50 & 72.48 $\pm$ 0.52 & 13.50 $\pm$ 1.14 \\
		& SechKAN & \textbf{72.97 $\pm$ 0.62} & \textbf{73.48 $\pm$ 0.41} & 9.85 $\pm$ 0.24 \\
            \hline
	\end{tabular}
	\label{tab:main_results}
\end{table*}

\begin{figure*}[!ht]
  \centering
  \includegraphics[scale=0.34]{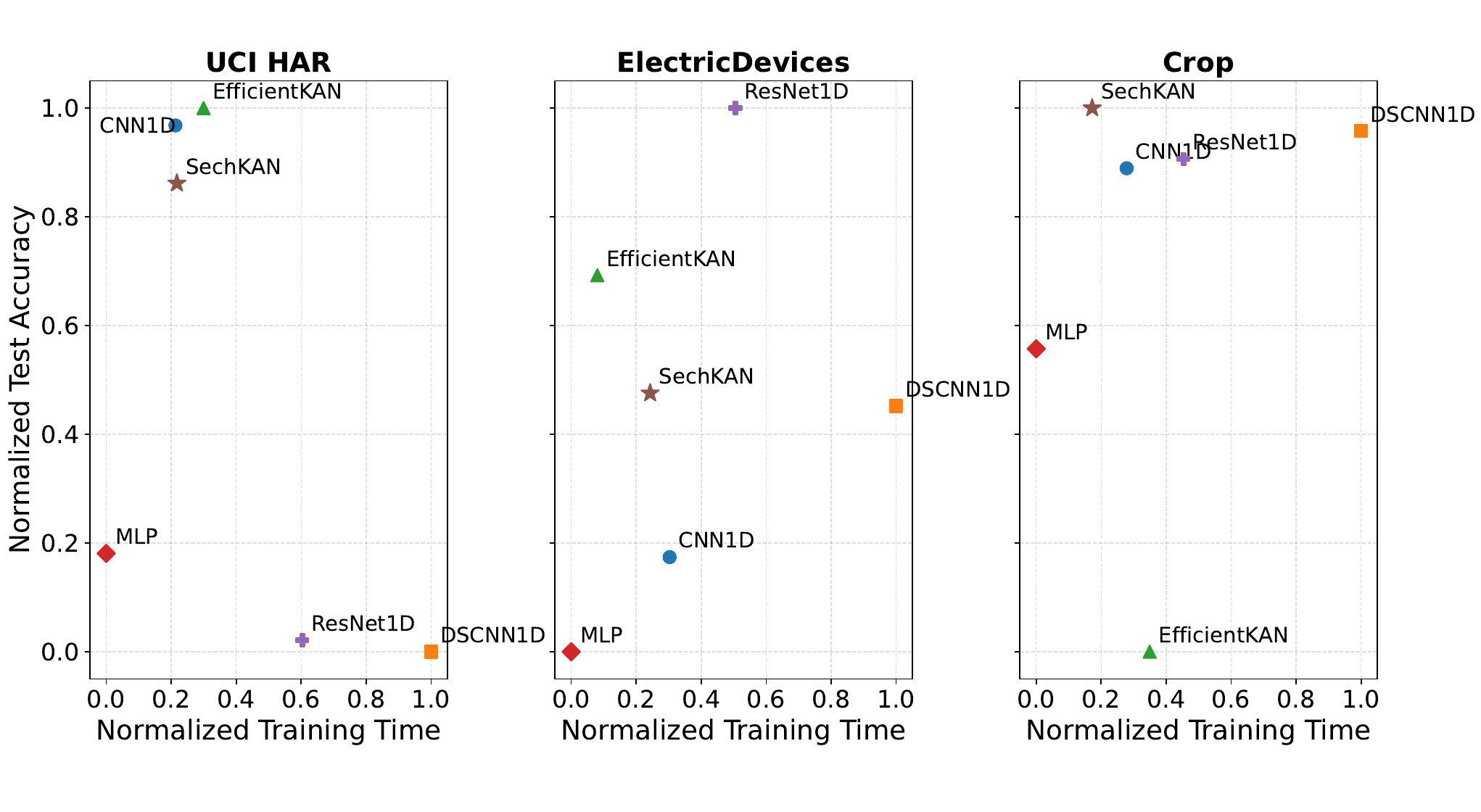}
  \caption{Normalized test accuracy versus training time for the evaluated models on (a) UCI HAR, (b) ElectricDevices, and (c) Crop. Both test accuracy and training time are min--max normalized to $[0,1]$ independently for each dataset to illustrate the trade-off between predictive performance and training efficiency.}
  \label{fig:training_eff}
\end{figure*}

~\Cref{tab:main_results} shows that SechKAN achieves competitive performance on the three datasets. On UCI HAR, SechKAN performs close to the best KAN-based model and is also comparable to CNN1D, while outperforming ResNet1D and DSCNN1D. On ElectricDevices, SechKAN achieves competitive performance, although ResNet1D obtains the highest test accuracy. SechKAN nevertheless outperforms CNN1D and MLP and remains comparable to DSCNN1D. On Crop, SechKAN achieves the highest test accuracy among the evaluated models, outperforming both DSCNN1D and ResNet1D. These results indicate that SechKAN provides competitive classification performance across datasets with different input dimensions and numbers of classes, with particularly strong performance on Crop.

~\Cref{fig:training_eff} complements these results by illustrating the relationship between test accuracy and training time. On UCI HAR, SechKAN combines high test accuracy with relatively low training time, whereas on ElectricDevices it provides an intermediate accuracy--time trade-off. On Crop, SechKAN achieves both the highest test accuracy and relatively low training time, placing it in a favorable region of the accuracy--time plot. In general, SechKAN achieves competitive accuracy without incurring the substantially higher training times observed for some deeper convolutional architectures, although its relative advantage depends on the dataset.

\subsection{Ablation studies}
\begin{figure*}[!ht]
  \centering
  \includegraphics[scale=0.6]{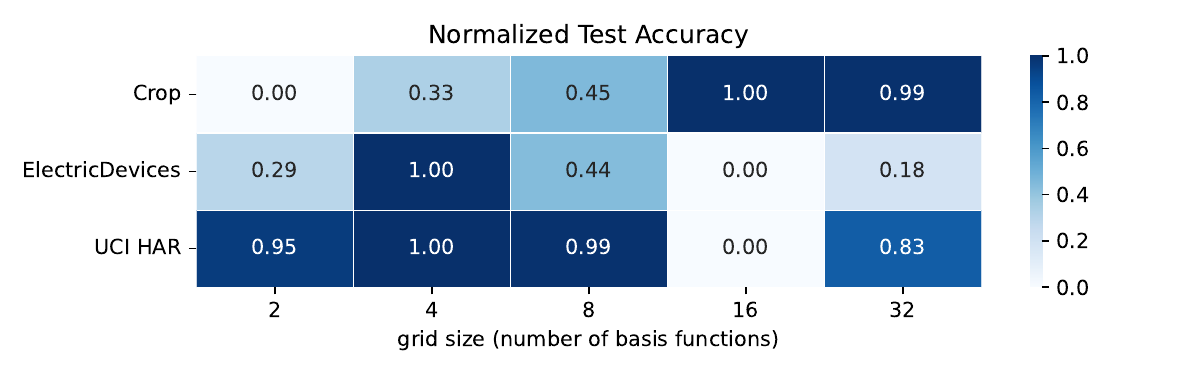}
 \caption{Ablation study of the number of basis functions in SechKAN using 2, 4, 8, 16, and 32 basis functions. All other configurations are kept identical to those used in the main experiments, with only the number of basis functions varied. The reported test accuracy values are averaged over three runs using seeds 0--2 and normalized to the range $[0,1]$ within each dataset, where 0 and 1 denote the worst and best results, respectively.}
  \label{fig:ab_grid}
\end{figure*}

\begin{figure*}[!ht]
  \centering
  \includegraphics[scale=0.6]{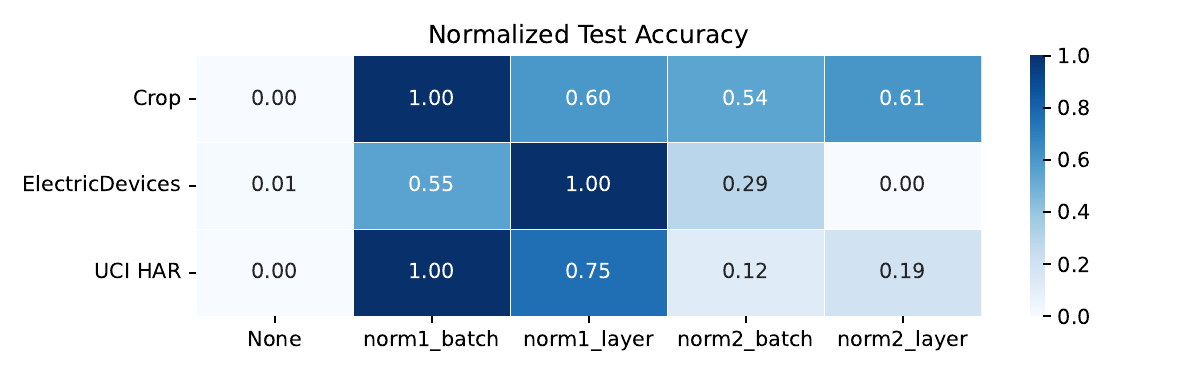}
\caption{Ablation study of normalization in SechKAN, comparing BatchNorm and LayerNorm applied at Norm1 and Norm2, as well as the configuration without normalization. All other configurations are kept identical to those used in the main experiments, with only the normalization type and position varied. The reported test accuracy values are averaged over three runs using seeds 0--2 and normalized to the range $[0,1]$ within each dataset, where 0 and 1 denote the worst and best results, respectively.}
  \label{fig:ab_norm}
\end{figure*}

In this section, we perform ablation studies on the grid size and normalization types (LayerNorm and BatchNorm) for SechKAN on three datasets. The training configuration of SechKAN is kept the same as in the main experiments, with only the grid size and normalization configuration changed. The ablations were conducted after the main comparison and were not
used to select the configurations reported in \Cref{tab:model_architectures}. The results in \Cref{fig:ab_grid} and \Cref{fig:ab_norm} show that SechKAN's performance is affected by both the grid size and the normalization configuration.

In \Cref{fig:ab_grid}, the effect of grid size varies across datasets. The highest normalized test accuracy is obtained with grid sizes of 16, 4, and 4 for Crop, ElectricDevices, and UCI HAR, respectively. This shows that increasing the grid size does not always improve performance. In \Cref{fig:ab_norm}, applying BatchNorm at Norm1 achieves the highest normalized test accuracy on all three datasets, while the configuration without normalization generally gives the lowest performance. LayerNorm at Norm1 also gives competitive results, whereas normalization at Norm2 gives less consistent results. These results show that grid size and normalization placement can affect SechKAN performance across different datasets.

However, these experiments cover only a limited set of configurations.
More extensive experiments on SechKAN's hyperparameters and
architectural choices are needed to better understand their effects.

\section{Limitations}
This study has several limitations. First, SechKAN was evaluated on only three 1D classification datasets. Although these datasets differ in input dimensions, sample sizes, and numbers of classes, they do not cover a wide range of domains. Therefore, the results may not generalize to other types of data, such as 2D and 3D data. Second, we focused only on classification and did not evaluate SechKAN on other tasks, such as regression and forecasting. Third, we did not perform an extensive search of SechKAN hyperparameters, including the activation function, learnable width, and skip connection. Therefore, the selected grid sizes and normalization settings may not be optimal for all datasets. Finally, the comparison included only one MLP, one KAN baseline, and three CNN-based models. More experiments with diverse datasets, tasks, hyperparameter settings, and baseline models are needed to better understand the generalizability and effectiveness of SechKAN.

\section{Conclusion}

In this paper, we investigated the effectiveness of SechKAN for one-dimensional classification by combining hyperbolic secant functions with a 1D projection. SechKAN was evaluated against EfficientKAN, MLP, CNN1D, ResNet1D, and DSCNN1D on three benchmark datasets: UCI HAR, ElectricDevices, and Crop, under approximately comparable parameter budgets. We also conducted ablation studies to examine the effects of grid size and normalization configuration. The results show that SechKAN achieves competitive performance across the three datasets, obtaining the highest test accuracy on Crop while maintaining relatively low training time. Its performance varies across datasets, indicating that the effectiveness of evaluated architectures depends on the characteristics of the one-dimensional classification task.

The ablation studies show that both grid size and normalization configuration affect SechKAN performance, with the best configurations varying across datasets. However, the evaluation is limited to three 1D classification datasets and a limited set of baseline models. Future work will extend the evaluation to a broader range of datasets and tasks, including regression and forecasting, and investigate a wider range of architectural designs and hyperparameter configurations.

\section*{Acknowledgements}
This research was funded by the University of Information Technology, Vietnam National University Ho Chi Minh City, under Grant No. S4-2027-80616.

%
%
%
\bibliographystyle{splncs04nat}
\bibliography{references}
\end{document}